\documentclass[twocolumn]{article}

\usepackage[a4paper, total={7in, 8in}]{geometry}
\usepackage[utf8]{inputenc}
\usepackage[english]{babel}

\usepackage{graphicx} 
\usepackage{multirow}
\usepackage{array}
\usepackage[dvipsnames]{xcolor}
\usepackage{booktabs}
\usepackage{hyperref}
\usepackage[font=tiny]{caption}
\usepackage{subcaption}
\usepackage{bm}

\usepackage{mathtools}
\usepackage{amsthm}
\usepackage{amssymb}
\usepackage{amsmath}

\usepackage[ruled,english,onelanguage]{algorithm2e}

\usepackage{parskip}
\usepackage{csquotes}
\usepackage[backend=biber]{biblatex}
\usepackage[title]{appendix}

\newtheorem{innercustomthm}{Lemma}
\newenvironment{customthm}[1]
  {\renewcommand\theinnercustomthm{#1}\innercustomthm}
  {\endinnercustomthm}
\newtheorem{theorem}{Theorem}
\newtheorem{lemma}[theorem]{Lemma}
\theoremstyle{definition}
\newtheorem{definition}[theorem]{Definition}
\theoremstyle{remark}

\newcommand\MyBox[2]{
  \fbox{\lower0.8cm
    \vbox to 2cm{\vfil
      \hbox to 2cm{\hfil\parbox{1.6cm}{#1\\#2}\hfil}
      \vfil}%
  }%
}

\newcommand*\samethanks[1][\value{footnote}]{\footnotemark[#1]}

\usepackage{authblk}

\title{Local Statistical Parity for the Estimation  of \\ Fair Decision Trees}

\author{
   Andrea Quintanilla\thanks{Computer Science Department, CIMAT} \thanks{University of Edinburgh}
  \and
  Johan Van Horebeek \samethanks[1]
}

\date{\today}

\begin{document}
\maketitle 

\begin{abstract}

Given the high computational complexity of decision tree estimation, classical methods construct a tree by adding one node at a time in a recursive way.  To facilitate promoting fairness, we propose a fairness criterion local to the tree nodes. 
We prove how it is related to the Statistical Parity criterion, popular in the Algorithmic Fairness literature, and show how to incorporate it into standard recursive tree estimation algorithms.

We present a tree estimation algorithm called Constrained Logistic Regression Tree (C-LRT), which is a modification of the standard CART algorithm using locally linear classifiers and imposing restrictions as done in Constrained Logistic Regression. 

Finally, we evaluate the performance of trees estimated with C-LRT on datasets commonly used in the Algorithmic Fairness literature, using various classification and fairness metrics.
The results confirm that C-LRT successfully allows to control and balance accuracy and fairness. 
\end{abstract}

\section{Introduction}
\label{sec:intro}
In Algorithmic Fairness (AF), the goal is to ensure that machine learning models used for decision-making related to individuals, such as resource allocation, do not unfairly discriminate or privilege individuals based on characteristics that are beyond their control or rooted in historical biases (e.g., ethnicity or gender) \cite{Kusner2017}, \cite{barocas-hardt-narayanan}.

This leads to two research questions \cite{Mehrabi2021}: (1) how to define a criterion that formalizes the notion of fairness \cite{Dwork2012}, \cite{Bilal2019}, and (2) how to construct a model that satisfies that criterion \cite{Johndrow2019}, \cite{Yurochkin2020}. In this paper, we focus on the latter and we aim to incorporate them in the model estimation stage \cite{Bilal2019}, \cite{Johndrow2019}.

Besides fairness, another crucial aspect in automated decisions related to individuals is transparency \cite{wing2021}, \cite{rudin2019}. It is essential that both the decision maker and the affected individual understand the reasons behind a decision when granting or denying a resource or imposing a sanction. Therefore, in this paper we use decision trees as basic classifiers.

The structure of the article is as follows: first, we review briefly previous work in the field of AF that is relevant to the proposed method and we introduce the notation and terminology that will be used.
Next, in Section \ref{sec:estimation_theory}, we propose a fairness criterion at the local node level and we show how local properties of decision nodes link to Statistical Parity of the global decision tree. Finally, in Section \ref{sec:c-lrt}, we propose and evaluate C-LRT, an algorithm for tree estimation that aims to promote fairness using the obtained theoretical relation between local and global parity.

\subsection{Relevant Work}
\label{sec:rel_work}






In the last decade several methods were proposed to incorporate fairness in decision trees, the majority focusing on  group fairness (\cite{Mitchell2021}, \cite{verma2018}) and only a few on individual fairness (e.g.  \cite{Vargo2021}, \cite{Ilvento2020}). 

A mathematically neat solution but with a limited practical utility was proposed in \cite{Aghaei2019}.  It offers a non-recursive approach, modeling decision trees using linear equations and imposing group fairness constraints as linear restrictions. The proposal has notable advantages. It ensures optimality and is highly relevant to our work because to our knowledge it is the only proposal  that ensures group fairness criteria without recurring to heuristics. However, the trees used in \cite{Aghaei2019} are still quite shallow due to the computational complexity of the optimization problem, which severely limits their predictive capacity. \medskip

Most often,  one looks for approximately solutions using heuristics in the tree construction process. An example of this approach  is  Dependency-Aware Tree Construction (IGC-IGS)\cite{Kamiran2010}, it served as a starting point of our proposal.   
IGC-IGS constructs trees recursively. To promote fairness, the choice of the  test or split in decision nodes, besides aiming to increase local accuracy in classification, incorporates fairness criteria in the selection process. 
In IGC-IGS, this is achieved by calculating the  information gain regarding the protected variable $A$ and the information gain regarding the output variable/classes $Y$. Although intuitively appealing, no formal justification is provided about the fairness of the final classifier.\medskip

 In our proposal, at each tree node, we will construct a classifier inspired by the Constrained  Logistic Regression Model (C-LR) introduced by \cite{Bilal2019}. In essence,  it estimates a logistic regression model and imposes fairness constraints by smoothing Statistical Parity controled by a parameter $c$. This smoothing results in a convex optimization problem, ensuring the existence of a Pareto minimum that simultaneously minimizes the log-likelihood and satisfies the fairness constraint. This minimum can be found using iterative methods such as Sequential Quadratic Programming (SQP) \cite{nocedal}. Empirically, $c$ proves to be a parameter that allows users to trade off between accuracy and fairness metrics. We emphasize that the exact election of the trade-off is not a mathematical problem but a social decision. Another advantage of the work by \cite{Bilal2019} over other AF methods is that in real-world problems, there are often multiple protected variables, and these variables typically have more than two values. Nevertheless, AF methods tend to limit themselves to a single binary variable. However, simple variations of C-LR can accommodate cases where $A$ represents more than one protected variable with multiple levels. \medskip

\subsection{Notation}
In the sequel, $Y$ represents a random variable for the ``output'' variable we aim to model. We assume it is binary, taking values of $-1$ and $1$, with $Y=1$ representing an advantageous outcome for the classified individual. The predictors define a random vector $X=(X_1,\ldots, X_p)$.  We assume that its components are numerical variables unless otherwise specified. We denote by $\mathcal{X}$ the input space, and by $\mathcal{Y}$ the output space. The random variable $A$ represents the so-called ``protected attributes''. These attributes aim to represent characteristics of individuals that should not be the basis for decisions. We assume that $A$ is not one of the predictors. $A$ is binary, with $A=0$ encoding the value associated with historically discriminated groups in the decision context and $A=1$ for others. We assume that we have a data set $D$ consisting of $n$ training samples, where $D$ is a sample of $(X,A,Y)$. The training data $D$ will be used to construct a model $\widehat{Y}(X)$ for $Y$. In our case,  $\widehat{Y}(X)$ will be a decision tree.

 A decision tree $T:\mathcal{X}\rightarrow \mathcal{Y}$ is a function that can be represented by nodes $t$ and relationships between them. We will focus on binary trees. To define the elements of $T$ and their evaluation method, let us consider a tree $T$ and one of its nodes $t$:

\begin{itemize}
    \item If $t$ is a decision node, it has associated a test of the form $I( X_{j_t} < r_t)$, where $I$ is the indicator function, $j_t \in \{1, \ldots,p \}$, $X_{j_t}$ is a predictor, and $r_t\in \mathbb{R}$.
    \item If $t$ is a decision node, we will denote its left and right children as $t_L$ and $t_R$, respectively. We will say that $t$ is the parent of $t_L$ and $t_R$.
    \item The root node of $T$ is the only node that has no parents. Occasionally, we will denote it as $t_0$.
    \item If $t$ is a terminal node, it has an associated label or constant prediction $\widehat{t} \in \{1, -1\}$.
    \item $\widetilde{T}$ will denote the collection of terminal nodes in $T$.
    \item Given a tree $T$, a case $x\in \mathcal{X}$ is evaluated in the nodes $t$ of the tree recursively, starting from the root node and ending at a single terminal node. The evaluation $t(x)$ is performed in two ways depending on whether node $t$ is a decision or terminal node:
        \begin{itemize}
        \item If $t$ is a decision node, the evaluation is $t(x) = I(x_{j_t} < r_t)$, where $j_t\in \{1,\ldots,p\}$, and $x_{j_t}$ is the $j_t$-th coordinate of case $x$. Depending on the true or false result of the test, it is assigned to the left or right child of $t$, respectively. The evaluation continues at the assigned node.
        \item If $t$ is a terminal node, $t(x)=\widehat{t}$, where $\widehat{t}$ is equal to the constant prediction associated with that terminal node. The prediction assigned to the case by the tree is $T(x)=t(x)=\widehat{t}$.
    \end{itemize}
\end{itemize}

\section{Local and Global Parity}
\label{sec:estimation_theory}
We introduce a local fairness criterion at the level of a node, associated with a global fairness criterion; the ultimate goal is that if the nodes meet the local criterion, the tree will satisfy the global fairness criterion as well.

To this end, we define first the global criterion we focus on and which is widely used in the AF literature \cite{LeQuy2022}.\medskip

\begin{definition}[Statistical Parity]\label{def:statistical_parity}
A model $\widehat{Y}(X)$ satisfies the Statistical Parity criterion if:
    $$
    P(\widehat{Y}(X) = 1 \mid A=0)=P(\widehat{Y}(X) = 1 \mid A=1)
    $$
    and
    $$
    P(\widehat{Y}(X) = -1 \mid A=0)=P(\widehat{Y}(X) = -1 \mid A=1)
    $$
    or, equivalently:
    $$\widehat{Y}(X)\perp A.$$

\noindent Intuitively, it aims to ensure that individuals  have equal opportunities to be classified in the advantaged class ($Y=1$) or the disadvantaged class ($Y=-1$) regardless of their protected group.
\end{definition}

\noindent For our proposed fairness criterion at the node level, the following definition is fundamental. It describes  for each case $x \in dom(T)$, at which nodes of $T$, $x$ will be evaluated in and at which nodes it won't.

\begin{definition} Let $T$ be a tree and $t$ a node of it.
\begin{itemize}
    \item We define the set $dom(T_t)$ as the domain of $T_t$. Intuitively, this represents the region of $\mathcal{X}$ associated with node $t$. We define it recursively: if $t$ is the root, $dom(T_t)$ is the same as the domain of $T$, i.e., $dom(T_t)=dom(T)=\mathcal{X}$. If $t$ is a child of a node $t'$:
    $$dom(T_t)
    = \left\{\begin{array}{lc}
        \{x\in dom(T_{t'}):  x_{j_{t'}}<r_{t'}\}, &  \text{if} \quad t=t'_{L}\\
        \{x\in dom(T_{t'}):  x_{j_{t'}}\geq r_{t'}\}, &  \text{if} \quad t=t'_{R}
    \end{array}\right.
    $$

    Given $x\in dom(T)$, with this definition, we identify the nodes $t$ in $T$ where $x$ is evaluated or ``falls'',
    
    \item $T_t:dom(T_t)\rightarrow \mathcal{Y}$ is known as the branch of $T$ rooted in $t$. It is also a decision tree and is composed of $t$ and its descendants in $T$. The way to evaluate a case $x \in dom(T_t)$ in tree $T_t$ is as explained in the previous section. Note that in the case where $t$ is a terminal node, $T_t=t$ is a single-node tree.
   
If $t$ is the root node of $T$, we denote $T_L:= T_{t_L}$ and $T_R:= T_{t_R}$. $T_L$ and $T_R$ are known as the primary left branch and the primary right branch of $T$, respectively.    
\item We define the function $dom_{T_t}: dom(T)\rightarrow \{0, 1\} $ as:
            $$dom_{T_t}(x)
            = \left\{\begin{array}{ll}
                0, & \quad \text{if} \quad x \notin dom(T_t)\\
                1, & \quad \text{if} \quad x \in dom(T_t)
            \end{array}\right.
            $$
    \end{itemize}
\end{definition}

We are now ready to propose the local equity criterion.

\begin{definition}[Local Statistical Parity]

If $t$ is a node of a tree $T$, we say that branch $T_t$ satisfies the Local Statistical Parity criterion with respect to $A$ if:
$$A \perp dom_{T_t}(X).$$
\end{definition}

\noindent We now formalize the relationship between Local Statistical Parity and Statistical Parity. The proof is straightforward and can be found as Lemma \ref{lema:tperpa_dem} in Appendix \ref{appen:dems}.

\begin{lemma}\label{lema:tperpa}
Given a tree $T$, if for every terminal node $t$, $T_t$ satisfies the Local Statistical Parity criterion, i.e., $A \perp dom_{T_t}(X)$, then $A\perp T(X)$.
\end{lemma}

As we mentioned, decision tree estimation algorithms are typically recursive: in each step, a node $t$ is added, and one decides whether it is a terminal or decision node. If $t$ is a decision node, a test $I(x_{j_t} < r_t)$ is chosen. In the rest of this section, we will prove a theorem that states that if, in a decision tree $T$, the tests associated with the decision nodes are ``independent of the protected variable'', then it ensures that the model $\bm{T(X)}$ complies with the Statistical Parity criterion. This will be very useful for incorporating equity criteria during estimation, and we will provide an algorithm as an example.\medskip

\noindent As independence between $A$ and the tests of the decision nodes $I(0 < \theta_t^T X_{j_t})$ is not always formally well defined because they might act over possible different spaces, with the following definition of $I_t^*$ functions, we extend the domain of the test functions $I(0 < \theta_t^T X_{j_t})$ to the same space as $A$ in such a way that, given $dom_{T_t}(X)=1$, $I_t^*(X) \equiv I(0 < \theta_t^T X_{j_t})$. 

\begin{definition}\label{def:ind_ext}
  Let $T$ be a decision tree, and $t$ be a decision node of $T$. Let $\alpha$ be fixed an arbitrary number in $\mathbb{R}\setminus \{0,1\}$. With the function $I^{*}_t(x):dom(T)\rightarrow \{0,1,\alpha\}$, we extend the domain of the test function associated with $t$. For each $x\in dom(T)$, it is defined as:
    $$I^{*}_t(x)
    =\left\{\begin{array}{lc}
        \alpha, & \quad \text{if} \quad x\notin dom(T_t)\\
        I(0 < \theta_t^T x_{j_t}), & \quad \text{if} \quad x\in dom(T_t)
     \end{array}\right.
    $$
Note that in the case where $t$ is the root node of $T$, $I^{*}_t(x)\equiv  I(0 < \theta_t^T x_{j_t})$.
\end{definition}\medskip

\noindent The following lemma states that if the tests associated with the decision nodes are independent of the protected variable, then all nodes in the tree satisfy the Local Statistical Parity criterion. The proof can be found as Lemma  \ref{lemma:c-lrt_dem} in Appendix \ref{appen:dems}.

\begin{lemma}\label{lema:c-lrt}
      Let $T$ be a decision tree, and $t_0$ be its root node. Assuming that the following holds for every decision node $t$ of $T$: $A \perp I_t^*(X)\mid dom_{T_t}(X)=1$, then every node (both decision and terminal) in $T$ satisfies the Local Statistical Parity criterion. That is:
      $$\forall t \text{ node of } T: \quad A \perp dom_{T_t}(X).$$
    
\end{lemma}

The following theorem is the main contribution of this paper, as it allows us to link theory with practice as shown in the next Section.

\begin{theorem}\label{teo:c-lrt}
      Let $T$ be a decision tree, and $t_0$ be its root node. If, for every decision node $t$ of $T$, $A \perp I_t^*(X)\mid dom_{T_t}(X)=1$, then $A\perp T(X)$.  
\end{theorem}

\noindent Note that Theorem \ref{teo:c-lrt} directly follows from Lemma \ref{lema:c-lrt} and Lemma \ref{lema:tperpa}.

\section{Constrained Logistic Regression Tree (C-LRT)}
\label{sec:c-lrt}

Exploiting Theorem \ref{teo:c-lrt}, we propose an estimation algorithm for decision trees that is very similar to the standard CART algorithm by \cite{Breiman1984}, with the exception that we incorporate fairness criteria when choosing decision tests at each node. To promote the independence of the tests $I^*_t$ associated with nodes $t$ from the variable $A$, we will use the fact that they are linear classifiers and adopt the C-LR algorithm by \cite{Bilal2019}. This algorithm provides a way to incorporate the Statistical Parity criterion into linear classifiers, such as Logistic Regression. To simplify notation, in the sequel we redefine $x$ as:
        $$x := (1, x^T)^T \in \mathbb{R}^{p+1}.$$

\noindent C-LR estimates a Logistic Regression model of the form:
$$
\widehat{Y}_{\theta} \left(x\right)=\operatorname{sign}(sd_{\theta}(x))
$$

\noindent where $\theta$ parametrizes the family of Logistic Regression functions, and $sd_\theta(x):=\theta^{T} x$ is the signed distance of $x$ to the decision boundary determined by $\theta$.

However, to estimate $\theta$ by minimizing the usual goodness-of-fit function $\ell$ of Logistic Regression, \cite{Bilal2019} impose a fairness constraint:
\begin{equation*}
\begin{array}{ll}
\theta = \arg \min_{\theta\in \mathbb{R}^{p+1}} & \ell(\theta, D)\\
\text { subject to } & \left| \widehat{Cov} \left(sd_{\theta}(X), A, D\right) \right|\leq c. 
\end{array}
\end{equation*}

\noindent where $\widehat{Cov}$ is the sample covariance, and $c>0$. In other words, they restrict the feasible area to those $\theta$ for which the estimated covariance between the signed distance and the protected variable is bounded by $c$. 


To incorporate fairness criteria during the tree construction based on the training set $D$, we will also need the following notation:
\begin{itemize}
    
    \item For each node $t$, we define:
    $$D_t=\{(x,a,y)\in D: x\in  \text{dom}(T_t)\}.$$
    
    $D_t$ represents the cases $(x,a,y)$ in $D$ for which $x$ falls into the region associated with node $t$.
    
    \item For each predictor $j=1, \ldots, p$, we define:
    $$D_{t, j}=\{(x_{j},a,y): (x,a,y) \in D_t \}$$ 
    $D_{t, j}$ represents the same tuples as in $D_t$, except that the predictors $x$ are projected onto their $j$-th coordinate.

\end{itemize}

\noindent We have now all ingredients to define the \textbf{Constrained Logistic Regression Tree (C-LRT)} algorithm. This algorithm is identical to a standard algorithm like CART\footnote{For a detailed description of CART, see \cite{Breiman1984} or \cite{Ripley2005}}, except for how it chooses tests $I(0 < \theta_t^T X_{j_t})$ for the decision nodes. In Algorithm \ref{alg:c-lrt_split}, we detail the method and call it Constrained Logistic Regression Tree Split (C-LRT split).

\begin{algorithm}[ht]
\DontPrintSemicolon
\caption{C-LRT split}
\SetKwInOut{Input}{Input}
\SetKwInOut{Output}{Output}
\Input{$D_t$: set of cases in node $t$; $c >0$ equity constraint.}
\Output{ $X_{j_t}, \theta_{j_t}$: optimal decision parameters $I(0<\theta_{j_t}^T X_{j_t})$ for $t$.}
\For{$j\leftarrow 1$ \KwTo $p$}{ 
    $D_{t, j} \longleftarrow \{\left( x_{j},a,y\right): (x,a,y) \in D_t \}$\;
    $\theta_j \longleftarrow  \arg \min_{\theta\in \mathbb{R}^2}  \ell(\theta, D_{t, j})$  subject to $\scriptstyle{\left| \widehat{Cov} \left(sd_{\theta}(X_{j_t}), A, D_{t, j}\right) \right|\leq c }$ \;
}
$j_t \longleftarrow \arg \min_{j\in \{1,\ldots,p\}} \sum_{(x,a,y)\in D_t} I(y \neq sign(\theta_j^T x_{j_t}))$\;
\Return $X_{j_t}, \theta_{j_t}$
\label{alg:c-lrt_split}
\end{algorithm}

\noindent The general idea of Algorithm \ref{alg:c-lrt_split} is to fit C-LR independently for each predictor $X_j$ in $D_t$:
$$\widehat{Y}_{\theta_{j}}(X)=\operatorname{sign}\left(sd_{\theta_{j}}\left[X_{j}\right]\right)$$
\noindent and to choose the test $I$ as the $\widehat{Y}_{\theta_{j_t}}$ that minimizes the error. The key to C-LR in including the independence of the tests with respect to $A$ is to note that for $\widehat{Y}_{\theta_{j_t}}$ to satisfy the Statistical Parity criterion:
$$\widehat{Y}_\theta(X) \perp A$$

\noindent it suffices that:
\begin{equation}
\label{restricion}
  sd_{\theta_{j_t}}\left[X_{j_t}\right] \perp A  
\end{equation}

\noindent Furthermore, as in \cite{Bilal2019}, to incorporate (\ref{restricion}), we relax it by fixing a small constant $c>0$ and adding the constraint to LR that $\theta_{j_t}$ is such that:
$$-c \leq Cov(sd_{\theta_{j_t}}\left[X_{j_t}\right], A) \leq c$$

\noindent C-LRT inherits the advantages of C-LR, as this relaxation implies that both constraints (inequalities) are convex functions with respect to the parameter $\theta_{j_t}$ and can be used with iterative methods that ensure convergence and optimality \cite{nocedal}.

\section{Results}
\label{sec:results}

In this section, we implement and empirically evaluate C-LRT, the algorithm based on the theoretical properties discussed in Section \ref{sec:estimation_theory}. To do so, we conduct experiments using various datasets commonly used in the AF literature \cite{LeQuy2022}.

\subsection{Methodology}
\label{sec:methodology}

Each of the experiments presented in this section was repeated 30 times: in each repetition, we randomly sampled from the dataset and split it into a training subset (70\%) and a testing subset (30\%). We constructed models on the training subset and recorded various aspects of model performance on the testing subset. The evaluated aspects can be categorized into two types: predictive power and fairness.\medskip

\noindent To evaluate the predictive power of each model, as done in \cite{LeQuy2022}, we can measure the rates associated with the confusion matrix in the usual way (over the entire testing set). However, we can also calculate these rates by considering only the subsets defined by the protected variable $A$: the protected group (prot) consisting of cases $(x,a,y)$ from the testing set where $a=0$, and the non-protected group (non-prot) consisting of cases $(x,a,y)$ with $a=1$.

\begin{table}[ht]
            \captionsetup{font={small}}
\begin{center}
\renewcommand\arraystretch{1.8}
\setlength\tabcolsep{0pt}
\begin{tabular}{c >{\bfseries}r @{\hspace{0.9em}}c @{\hspace{0.9em}}c}
  \multirow{10}{*}{\parbox{0.5cm}{\bfseries $Y$}} & & \multicolumn{2}{c}{\bfseries $\widehat{Y}$}\\
  & & \bfseries 1 & \bfseries -1 \\
  & 1 & \MyBox{\small{True Positive (TP)}}{\small{$TP_{\operatorname{prot}}$ $TP_{\operatorname{non-prot}}$}} & \MyBox{\small{False Negative (FN)}}{\small{$FN_{\operatorname{prot}}$ $FN_{\operatorname{non-prot}}$}} \\[2.6em]
  & -1 & \MyBox{\small{False Positive (FP)}}{\small{$FP_{\operatorname{prot}}$ $FP_{\operatorname{non-prot}}$}} & \MyBox{\small{True Negative (TN)}}{\small{$TN_{\operatorname{prot}}$ $TN_{\operatorname{non-prot}}$}} 
\end{tabular}
\end{center}
    \caption{\label{tab:confusion_matrix} Confusion matrix for a classifier $\widehat{Y}$. TP, FN, FP, and TN are the usual counts over the entire testing set, while the counts with subscripts prot and non-prot are the analogous counts but restricted to the protected and non-protected groups, respectively.}
\end{table}

\noindent Since the datasets we work with are often imbalanced, we are interested in the \textit{True negative rate (TNR)} and \textit{True positive rate (TPR)}. Each of the aforementioned rates, $T$, can be calculated by considering only the protected group or the non-protected group, denoted as $T_{\text{prot}}$ and $T_{\text{non-prot}}$, respectively. For example, the TPR and TNR for the protected group are defined as follows:
    $$
    \mathrm{TPR}_{\text{prot }}=\frac{\mathrm{TP}_{\text{prot}}}{\mathrm{TP}_{\text{prot}}+\mathrm{FN}_{\text{prot}}}
    $$
    
    $$
    \mathrm{TNR}_{\text{prot }}=\frac{\mathrm{TN}_{\text{prot }}}{\mathrm{TN}_{\text{prot}}+\mathrm{FP}_{\text{prot}}}
    $$

\noindent Similarly, we can define $\mathrm{TPR}_{\text{non-prot}}$ and $\mathrm{TNR}_{\text{non-prot}}$ for the non-protected group.\medskip

\noindent To evaluate the predictive power of each model, we use the metrics \textit{accuracy} and \textit{balanced accuracy} in both groups of interest:
\begin{itemize}
    \item Accuracy:
    $$
    \text{Accuracy}=\frac{\mathrm{TP}+\mathrm{TN}}{\mathrm{TP}+\mathrm{TN}+\mathrm{FP}+\mathrm{FN}} 
    $$
    
    \item Balanced Accuracy (BA):
    $$
    \text{BA}=\frac{1}{2} \times\left(TPR +TNR \right)
    $$
    
    \item Balanced Protected Accuracy (BPA) in the protected group:
    \begin{align*}
    \text{BPA}=
    \frac{1}{2} \times\left(TPR_{\text{prot}} +TNR_{\text{prot}} \right)
    \end{align*}

    \item Balanced Non-Protected Accuracy (BNPA) in the non-protected group:
    \begin{align*}
    \text{BNPA}=
    \frac{1}{2} \times\left(TPR_{\text{non-prot}} +TNR_{\text{non-prot}} \right)
    \end{align*}
\end{itemize}

To assess the Statistical Parity of the resulting decision trees, we considered the following fairness metrics: Statistical Parity Difference (SP), p-rule, and n-rule. SP is defined as:
\begin{equation*}\label{eq:sp}
{\ \mathrm{SP}(\widehat{Y})=P(\widehat{Y}(X) = 1\mid A=1)-P(\widehat{Y}(X) = 1\mid A=0)}
\end{equation*}

\noindent SP can take values in the range $[-1, 1]$, where $\mathrm{SP}(\widehat{Y})=0$ indicates no discrimination with respect to $A$. A positive value of $\mathrm{SP}(\widehat{Y})$ within the range $(0,1]$ suggests discrimination against the protected group, while a negative value within the range $[-1,0)$ indicates discrimination against the non-protected group. This metric was proposed in \cite{Dwork2012} and is used in various works \cite{LeQuy2022}.\medskip

\noindent We also use the p-rule metric, which we define as follows. Let $y=1$. In cases where ${P\left(\widehat{Y}(X) = y \mid A=a\right)\neq 0}$ for $a=0,1$, we define:
\begin{multline*}
{ \operatorname{p-rule}(\widehat{Y}) =} \\{ \min \left(\frac{P\left(\widehat{Y}(X) = y \mid A=1\right)}{P\left(\widehat{Y}(X) = y \mid A=0\right)}, \frac{P\left(\widehat{Y}(X) = y \mid A=0\right)}{P\left(\widehat{Y}(X) = y \mid A=1\right)}\right) }
\end{multline*}

\noindent In cases where either denominator is zero, it indicates that the classifier is constant $\widehat{Y}(X) \equiv -1$, achieving Statistical Parity criteria in a less satisfactory manner. In such cases, we define:
$$\operatorname{p-rule}(\widehat{Y}) = 1 \quad \text{when}\quad \widehat{Y}(X) \equiv -1$$

\noindent If $y=-1$, we obtain an analogous metric $\operatorname{n-rule}$ in cases where ${ P\left(\widehat{Y}(X) = -1 \mid A=a\right)\neq 0}$ for $a=0,1$. We also define:
$$\operatorname{n-rule}(\widehat{Y}) = 1 \quad \text{when}\quad \widehat{Y}(X) \equiv 1$$

\noindent Note that the closer the values of $\operatorname{p-rule}(\widehat{Y})$ and $\operatorname{n-rule}(\widehat{Y})$ are to 1, the closer they are to satisfying the equalities involved in Definition \ref{def:statistical_parity} for $\widehat{Y}(X)=1$ and $\widehat{Y}(X)=-1$, respectively.\medskip

\noindent p-rule was proposed in \cite{biddle2017} and is used, for example, in \cite{Bilal2019}. We introduced the analogous n-rule metric as it, in conjunction with p-rule, allows us to quantify in a more complete way the Statistical Parity criterion.

\subsubsection{Datasets}

The datasets used for the evaluations are part of a list presented in the study \cite{LeQuy2022}. This list consists of publicly available datasets that have been used in at least three relevant articles in the field. In the same study, these datasets are classified into four major domains: financial, criminological, educational, and social welfare. Since our proposal (C-LRT) is focused on promoting Statistical Parity, we chose to evaluate it within each domain by selecting a dataset where a logistic regression (LR) model discriminates more against the protected group according to the SP metric. To do this, we replicated an experiment proposed in \cite{LeQuy2022}. In all cases, we used the attributes most frequently selected in the literature as protected attributes, as indicated by \cite{LeQuy2022}. Below, we describe the selected datasets. \medskip

\begin{itemize}
    \item \textbf{Adult.} This dataset was constructed from a census conducted in the USA in 1994 and consists of 48,842 cases. It is one of the most popular datasets in the field of Equal Opportunity in Machine Learning (EOML) and is available at \href{https://archive.ics.uci.edu/ml/datasets/adult}{adult}. The variable $Y$ is an indicator of whether a person earns more than \$50,000 USD annually or not, and $X$ consists of 13 demographic measures. Gender is fixed as the protected variable $A$ (with only male and female classes).
    
    \begin{table}[ht]
            \captionsetup{font={small}}
        \begin{center}
        \begin{tabular}{|>{\centering}p{0.10\linewidth}|c|c|} \hline
         $\quad\quad A$ \newline $\quad Y$ & \textbf{Female} & \textbf{Male} \\ \hline
        $\leq 50K$ & 13026 (88.6\%) & 20988 (68.8\%) \\ \hline
        $> 50K$ & 1669 (11.4\%) & 9539 (31.2\%) \\ \hline
        \end{tabular}
        \end{center}
        \caption{\label{tab:cont_adult} Contingency table between $Y$ and $A$ for Adult. Percentages were calculated by column.}
    \end{table}

    \item \textbf{COMPAS.} This dataset was created for a Pro\-Publica investigation on the algorithm ``Correctional Offender Management Profiling for Alternative Sanctions (COMPAS)'', which judges use to assist in decisions regarding parole \cite{angwin2016}. The dataset has been used in various studies on recidivism \cite{LeQuy2022} and contains 7,214 instances. It is available at \href{https://github.com/propublica/compas-analysis}{compas}. Each case corresponds to a person granted parole. The variable $Y$ indicates whether the individual reoffended after parole, and the predictors $X$ include demographic information, criminal history, and COMPAS-generated scores. Race is used as the protected variable $A$.
    
    \begin{table}[ht]
        \captionsetup{font={small}}
        \begin{center}
        \begin{tabular}{|>{\centering}p{0.20\linewidth}|c|c|} \hline
        $\quad\quad\text{A}$ \newline $\text{Y}$ & \textbf{African-American} & \textbf{Caucasian} \\ \hline
        $0$ (no recidivism) & 1514 (47.7\%) & 1281 (60.9\%) \\ \hline
        $1$ (recidivism) & 1661 (52.3\%) & 822 (39.1\%) \\ \hline
        \end{tabular}
        \end{center}
        \caption{\label{tab:cont_compas} Contingency table between $Y$ and $A$ for COMPAS. Percentages were calculated by column.}
    \end{table}

    \item \textbf{Ricci.} This dataset was generated for a case study handled in the Supreme Court of the USA \cite{ricci2009}. The case took place in a city in Connecticut, USA, in 2003, where exams were used to identify qualified firefighters for job promotions. It is a relatively small dataset consisting of 118 instances, and it is available at \href{https://www.key2stats.com/data-set/view/690}{ricci}. Each case corresponds to one of the aspiring firefighters. The variable $Y$ indicates whether they were promoted, and race is used as the protected variable $A$. The predictors $X$ consist of information about their performance on two exams and the position they aspire to (lieutenant or captain).
    
    \begin{table}[ht]
        \captionsetup{font={small}}
        \begin{center}
        \begin{tabular}{|>{\centering}p{0.20\linewidth}|c|c|} \hline
        $\quad\quad\text{A}$ \newline $\text{Y}$ & \textbf{Non-White} & \textbf{White} \\ \hline
        \textbf{False} (not promoted) & 35 (70.0\%) & 27 (39.7\%) \\ \hline
        \textbf{True} (promoted) & 15 (30.0\%) & 41 (60.3\%) \\ \hline
        \end{tabular}
        \end{center}
        \caption{\label{tab:cont_ricci} Contingency table between $Y$ and $A$ for Ricci. Percentages were calculated by column.}
    \end{table}

    \item \textbf{Law School.} This dataset was constructed for a longitudinal study, ``The Law School Admission Council National Longitudinal Study'', with information collected from surveys conducted at 163 schools in the USA in 1991 \cite{wightman1998}. It consists of 20,798 instances associated with different students and is available at \href{https://github.com/tailequy/fairness_dataset/tree/main/Law_school}{law school}. It is often used to predict whether a student will pass an entrance exam to a school on their first attempt or to predict their overall first-year average at the institution. In our case, we set the variable indicating whether the student passed the exam or not as the output variable $Y$. Race is used as the protected variable $A$, and the predictors $X$ include school reports and student questionnaires, containing demographic and educational information.
    
    \begin{table}[ht]
        \captionsetup{font={small}}
        \begin{center}
        \begin{tabular}{|>{\centering}p{0.10\linewidth}|c|c|} \hline
         $\quad\quad\text{A}$ \newline $\quad\text{Y}$ & \textbf{Non-White} & \textbf{White} \\ \hline
        $0$ & 916 (27.7\%) & 1377 (7.9\%) \\ \hline
        $1$ & 2391 (72.3\%) & 16114 (92.1\%) \\ \hline
        \end{tabular}
        \end{center}
        \caption{\label{tab:cont_law} Contingency table between $Y$ and $A$ for Law School. Percentages were calculated by column.}
    \end{table}
\end{itemize}

\subsection{Evaluations}

In this section, we will present and discuss the evaluations of the results of C-LRT. We will focus on studying the effects of the parameters $c$ and $\lambda$ on its performance.

\begin{figure*}[h]
    \centering
    \begin{subfigure}[b]{\textwidth}
        \centering
        \includegraphics[width=0.4\linewidth]{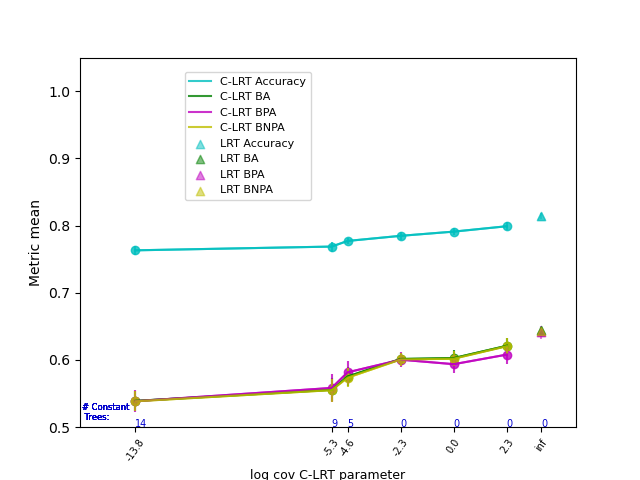}
        \includegraphics[width=0.4\linewidth]{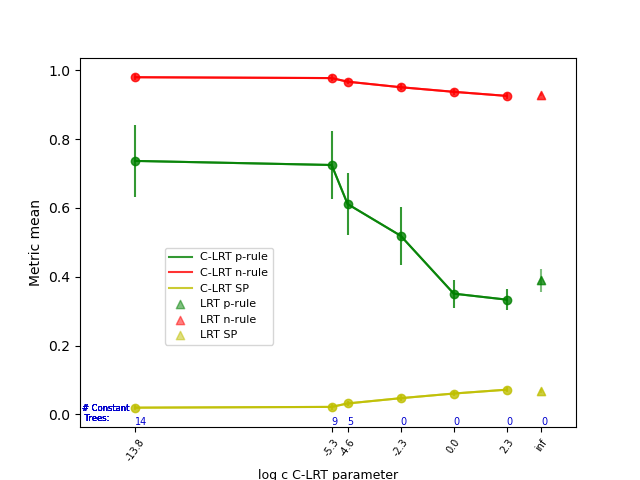}
        \caption*{(a) Adult.}
    \end{subfigure}
    
    \vskip\baselineskip
    
    \begin{subfigure}[b]{\textwidth}
        \centering
        \includegraphics[width=0.4\linewidth]{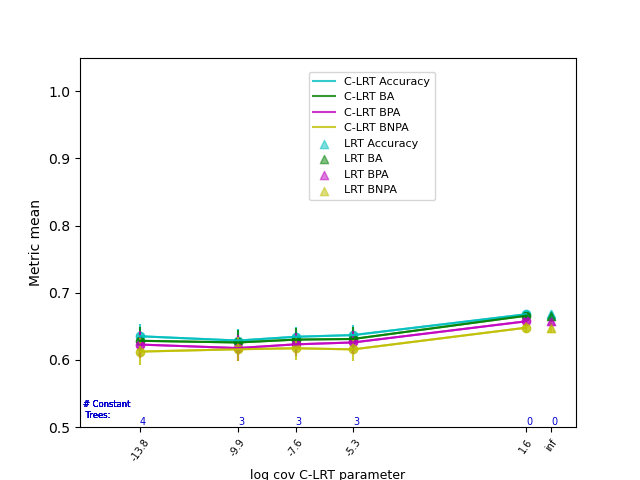}
        \includegraphics[width=0.4\linewidth]{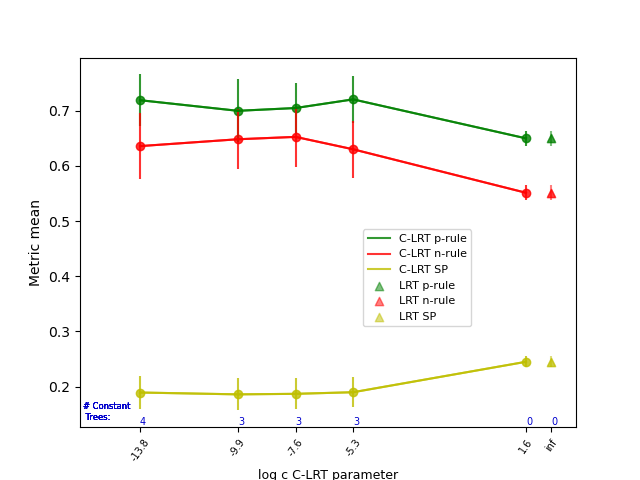}
        \caption*{(b) COMPAS.}
    \end{subfigure}
    
     \vskip\baselineskip
    
    \begin{subfigure}[b]{\textwidth}
        \centering
        \includegraphics[width=0.4\linewidth]{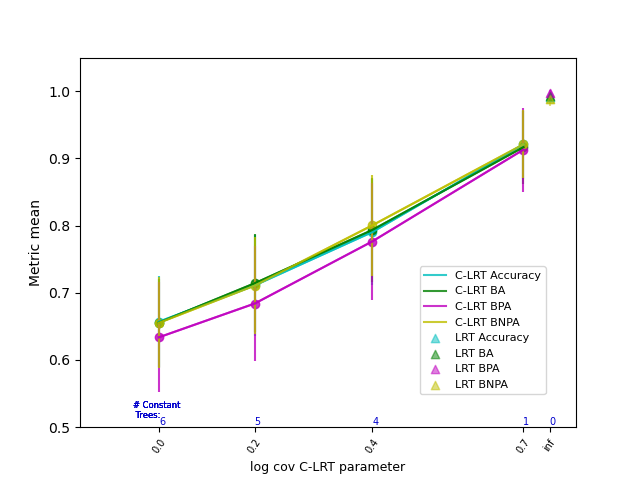}
        \includegraphics[width=0.4\linewidth]{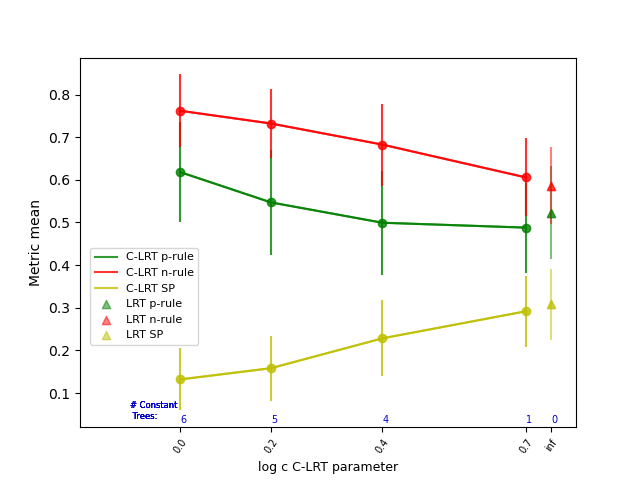}
        \caption*{(c) Ricci.}
    \end{subfigure}

     \vskip\baselineskip
    
    \begin{subfigure}[b]{\textwidth}
        \centering
        \includegraphics[width=0.4\linewidth]{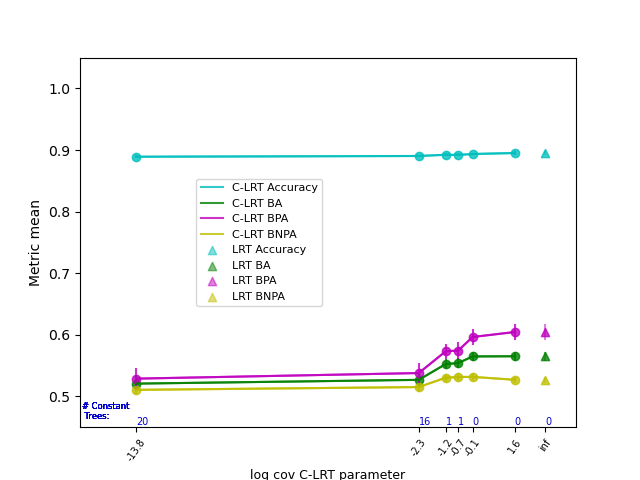}
        \includegraphics[width=0.4\linewidth]{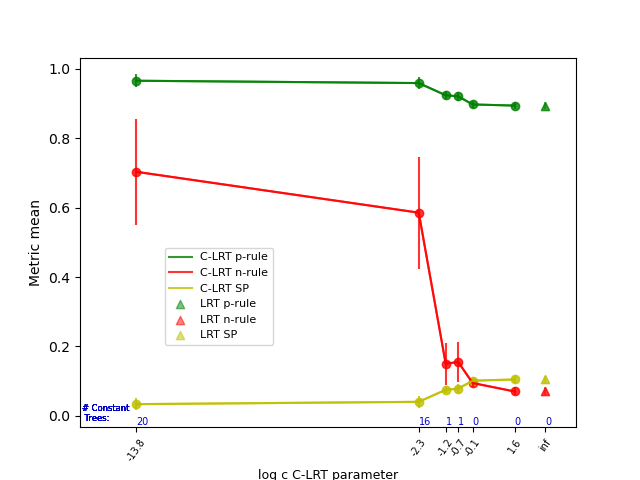}
    \caption*{(d) Law School.}
    \end{subfigure}
    
    \caption{Prediction metrics (left column) and fairness metrics (right column) for LRT and C-LRT. The x-axis is associated with the value of the parameter $c$ of C-LRT; for LRT, there is no $c$, represented as ``inf''. Each color corresponds to a metric, and the y-axis is the average of the metric over 30 experiments. The blue numbers on the x-axis indicate the count of constant decision trees associated with the value of $c$. Error bars represent the confidence interval of the average.}

    \label{fig:metrics}
\end{figure*}

Note that in Algorithm \ref{alg:c-lrt_split}, small values of the parameter $c$ correspond to stronger constraints, while larger values correspond to weaker constraints. In the graphs in Figure \ref{fig:metrics}, we display evaluations of C-LRT for different values of $c$. We also include evaluations of an algorithm for constructing decision trees identical to C-LRT, except that it removes fairness constraints when choosing node decisions. We denote this algorithm as ``Logistic Regression Tree (LRT)''.\medskip

We observe two expected phenomena: a) an inverse relationship (trade-off) between fairness measures and accuracy, meaning that as fairness measures improve, accuracy decreases, and vice versa; and b) when the value of $c$ is sufficiently large, the behavior of C-LRT is very similar to that of LRT.\medskip

Despite the fact that the constraints are applied at the local level, it is observed that the obtained decision trees  favor fairness and are controlled by the value of $c$. Stronger constraints yield better results in fairness measures. This suggests that the proposed method, although it works with a relaxation of the independence requirement of Theorem \ref{teo:c-lrt} in the trees it constructs, does promote improvements in the selected fairness metrics for capturing Statistical Parity. It is also noted that where the constraints are stronger, the highest averages in fairness metrics are achieved. However, in those same cases, there are relatively higher variances.\medskip

Furthermore, in Figure \ref{fig:metrics}, we also record the number of constant decision trees (on the x-axis). These trees consist of a single terminal node that maps all evaluated cases to the same class (the majority class). We observe that when constraints are stronger, these types of trees are selected more frequently to meet the constraints. However, there are values of $c$ where this strategy is rarely used, resulting in better accuracy values and significantly better fairness performance compared to the unconstrained tree, LRT.\medskip

\section{Conclusions and Future Work}

We introduced the concept of Local Statistical Parity and proved Lemma \ref{lema:tperpa}, which presents a sufficient local property in terms of Local Statistical Fairness for a decision tree to satisfy the criterion of Statistical Parity (globally).\medskip

The advantage of our theoretical contributions lies in their ability to modify recursive construction algorithms to promote fairness while preserving their recursive structure, thus avoiding the need for computationally intensive methods.\medskip

To put this into practice, we proposed C-LRT, an algorithm for decision tree construction that extends the method of algorithmic fairness, C-LR, centered on Logistic Regression models. To achieve this extension, we demonstrated Theorem \ref{teo:c-lrt}, which establishes that for a tree to satisfy the Statistical Parity criterion, it is sufficient for each decision node's split to satisfy the same criterion.\medskip

Another advantage of C-LRT (similar to standard decision trees) is that adjustments are made at each node in variable $X_j$, which can be advantageous in high dimensions. On the other hand, in C-LR and C-LRT, a relaxation is introduced: instead of requiring independence ($A \perp \widehat{Y}(X)$), it only imposes that the magnitude of covariance is less than a certain value specified through a regularization parameter. This parameter allows for a trade-off between accuracy and fairness.\medskip

In the results obtained with C-LRT, we observed that this parameter indeed allows the model's user to decide how to weigh accuracy versus fairness. Therefore, a contribution of this work is the successful adaptation of C-LR to decision trees. Of course, whether a classifier based on logistic regression or decision trees performs better depends on the nature of the data. A side effect observed multiple times with C-LRT is that when giving too much weight to fairness, the decision tree reduces to a single-node constant classifier.

\subsection*{Future Work}

It would be interesting to work with the fairness criterion directly instead of using a relaxation, i.e., calculating it for all possible splits and only considering decisions that meet it. Such calculations could be computationally feasible since CART calculates an impurity measure for all possible splits when choosing the associated decision for a node.\medskip

Another interest is to propose node-level (local) fairness criteria that have theoretical relationships with other fairness criteria from the literature.\medskip

Additionally, as mentioned earlier, it would be informative to conduct experiments on high-dimensional datasets.\medskip

Finally, it is important to investigate whether there are alternatives to reduce the occurrence of degenerate trees when imposing strict constraints.

\begin{appendices}

\section{Proofs}
\label{appen:dems}

\begin{customthm}{5}\label{lema:tperpa_dem}
Given a tree $T$, if for every terminal node $t\in \widetilde{T}$, $T_t$ satisfies the Local Statistical Parity criterion, i.e., $A \perp dom_{T_t}(X)$, then $A\perp T(X)$.
\end{customthm}
\begin{proof}
We must prove that:
\begin{align*}
     \forall\hat{y} \in \{-1,1\}, \forall a \in \{0,1\}: &  \\
     & P(T(X)=\hat{y}, A=a) =\\
     & P(T(X)=\hat{y}) P(A=a)
\end{align*}

Let $a \in \{0,1\}$ and $\hat{y} \in \{-1,1\}$. We know that $\mathcal{D} = \{dom(t): t \in \widetilde{T}\}$ forms a partition of $dom(T)$. Then, if $x\in dom(T)$:
$$T(x)=\hat{y} \leftrightarrow x\in \bigcup_{t\in \widetilde{T}: \hat{t}=\hat{y}}dom(t)$$

where, as a reminder, $\hat{t}$ is the prediction of $t$ for $t\in \widetilde{T}$. Also, recall that:
$$\text{if } t\in T: x\in dom(t) \leftrightarrow dom_{T_t}(x) = 1 $$

Therefore, using the fact that the elements of $\mathcal{D}$ are pairwise disjoint and the hypothesis, we conclude that:

\begin{flalign*}
      P(T(X)=\hat{y}, A=a) =& \\
      P(\bigvee_{t\in \widetilde{T}: \hat{t}=\hat{y}} dom_{T_t}(X)=1 \wedge A=a)=  \\
     \sum_{t\in \widetilde{T}: \hat{t}=\hat{y}} P(dom_{T_t}(X)=1, A=a) =\\
     \sum_{t\in \widetilde{T}: \hat{t}=\hat{y}} P(dom_{T_t}(X)=1) P(A=a) =\\
     P(\bigvee_{t\in \widetilde{T}: \hat{t}=\hat{y}} dom_{T_t}(X)=1) P(A=a) =\\
     P(T(X)=\hat{y}) P(A=a).\\
\end{flalign*}
\end{proof}

\begin{customthm} {7}\label{lemma:c-lrt_dem}
Let $T$ be a decision tree and $t_0$ its root node. Assuming that the following holds:
$$\forall t \text{ decision node of } T: A \perp I_t^*(X)\mid dom_{T_t}(X)=1,$$
then every node (decision and terminal) of $T$ satisfies the Local Statistical Parity criterion. In other words:
$$\forall t \text{ node  of } T: \quad A \perp dom_{T_t}(X).$$    
\end{customthm}  

\begin{proof}
Let $T$ be a tree as in the hypothesis. This time, we will proceed recursively but starting from the top of the tree. We will prove for each node $t$ that:
\begin{itemize}
    \item[a)] $A \perp dom_{T_t}(X)$
    \item[b)] if $t$ is a decision node, then $A \perp I^{*}_t(X)$.
\end{itemize}
Note that a) is the property we want to prove. On the other hand, b) is a property that will serve later as an auxiliary when proving a) for the children of $t$. Also note that property b) is similar to the hypothesis but involving an unconditional independence.\medskip
  
\textbf{Base Case.} Suppose that $t=t_0$ is the root of $T$. a) is true because $dom_{T_{t}}(X)\equiv 1$ is degenerate, so $A \perp dom_{T_{t}}(X)$. b) is also true because by hypothesis $A\perp I_{t}^{*}(X)\mid dom_{T_t}(X)=1$, and since $dom_{T_{t}}(X)$ is degenerated, this implies $A\perp I_{t}^{*}(X)$.\medskip

\textbf{Recursion.} 
Suppose that $t$ is a decision node for which a) and b) hold. We will now prove that properties a) and b) are true for its children, $t_L$ and $t_R$.
\begin{itemize}
    \item[a)] We must prove that:
    \begin{align*}
        \forall s \in \{t_L, t_R\} \forall a \in \{0, 1\} \forall d \in \{0, 1\}: \\
        P(A=a \mid dom_{T_s}(X)=d ) = P( A=a )
    \end{align*}

    Let $a\in \{0, 1\}$.\medskip
    \begin{itemize}
        \item Let's start with the case where $d=1$. Recall that for any $x \in dom(T_t)$, if its evaluation in $I(0<\theta_t^T x_{j_t})$ is positive, then $x$ is assigned to the left child of the node, and to the right child otherwise. Thus, we have:
        $$
            dom_{T_{t_L}}(X)=1 \text{ if and only if } I_{t}^{*}(X) = 1
        $$
    and:
    $$
        dom_{T_{t_R}}(X)=1 \text{ if and only if }
        I_{t}^{*}(X) = 0.
    $$

    Furthermore, since $t$ satisfies property b), we obtain:
    \begin{align*}
        P(A=a \mid dom_{T_{t_L}}(X)=1) = \\
         P\left(A=a \mid I_{t}^{*}(X) = 1 \right) = \\
          P(A=a)
    \end{align*}
    \begin{align*}
        P(A=a \mid dom_{T_{t_R}}(X)=1) = \\P\left(A=a \mid I_{t}^{*}(X) = 0 \right) = \\ P(A=a)
    \end{align*}

    \item For the case $d=0$, let's prove it for $s=t_L$ only, as the proof for $t_R$ is analogous. Let $x\in dom(T)$. Note that $dom_{T_{t_L}}(x) = 0$ happens if and only if: i) $x \in dom(T_{t_R})$, or ii) $x \notin dom(T_{t})$. As i) and ii) are mutually exclusive events, we have:
    \begin{flalign*}
             P(dom_{T_{t_L}}(X)=0, A=a) = \\
             P\left( \left[dom_{T_{t_R}}(X)=1 \wedge A=a \right] \right . \vee \\
             \left . \left[dom_{T_t}(X)=0 \wedge A=a \right] \right) = \\
             P\left( dom_{T_{t_R}}(X)=1, A=a \right) +  \\  
             P\left(dom_{T_t}(X)=0, A=a \right) = \\
             P(I_{t}^{*}(X)=0, A=a) + \\ 
             P\left(I_{t}^{*}(X)=\alpha, A=a \right).
    \end{flalign*}

     And, once again, using that $t$ satisfies property b):
     \begin{flalign*}
          P(dom_{T_{t_L}}(X)=0, A=a) = \\ 
          P\left(I_{t}^{*}(X)=0\right) P\left( A=a \right) + \\
          P\left(I_{t}^{*}(X)=\alpha\right) P\left( A=a \right) = \\
         \left[ P\left( I_{t}^{*}(X)=0\right) \right . + \\
         \left . P\left( I_{t}^{*}(X)=\alpha\right)\right]  P\left( A=a \right) = \\
         \left[ P\left( I_{t}^{*}(X)=0 \vee I_{t}^{*}(X)=\alpha\right)\right]  P\left( A=a \right) = \\
          P\left( dom_{T_{t_L}}(X)=0 \right) P\left( A=a \right).        
     \end{flalign*}

    \end{itemize}     
    \item[b)] To prove that $A \perp I^{*}_t(X)$, we will limit ourselves to the case of $t_L$, as the proof for $t_R$ is analogous. Let's verify that:    
        \begin{align*}
           \forall i\in \{0, 1, \alpha\}\quad \forall a\in \{0, 1\}: \\ P(I^{*}_{t_L}(X)=i, A=a) = P(I^{*}_{t_L}(X)=i)P(A=a).         
        \end{align*}

        Let $a\in \{0, 1\}$. For the case where $i\in \{0, 1\}$, observe that $I^{*}_{t_L}(X)=i$ implies that $dom_{T_{t_L}}(X)=1$, and therefore $dom_{T_{t}}(X)=1$. Thus:
        \begin{flalign*}
        P(I^{*}_{t_L}(X)=i, A=a) = \\
        P(I^{*}_{t_L}(X)=i, A=a, dom_{T_t}(X)=1) = \\
        P(I^{*}_{t_L}(X)=i, A=a \mid dom_{T_{t_L}}(X)=1) \\
        P(dom_{T_{t_L}}(X)=1)
        \end{flalign*}

        Now, using the hypothesis, we know that:
            \begin{flalign*}
                 P(I^{*}_{t_L}(X)=i, A=a) = \\
                P(I^{*}_{t_L}(X)=i\mid dom_{T_{t_L}}(X)=1) \\
                P( A=a \mid dom_{T_{t_L}}(X)=1) \\
                P(dom_{T_{t_L}}(X)=1)
            \end{flalign*}
        
        Also, it's important to note that, as we proved in the previous part, $t_L$ satisfies property a), which implies $P(A=a \mid dom_{T_{t_L}}(X)=1)=P(A=a)$. Hence, using the definition of conditional probability, $P(I^{*}_{t_L}(X)=i, A=a)$ becomes:
        \begin{align*}
          P(I^{*}_{t_L}(X)=i, A=a) = \\
          P(A=a) P(I^{*}_{t_L}(X)=i, dom_{T_t}(X)=1).  
        \end{align*}
            
        Finally, using again that $I^{*}_{t_L}(X)\in \{0, 1\}$ implies $dom_{T_t}(X)=1$, we conclude that:
            $$P(I^{*}_{t_L}(X)=i, A=a) = P(I^{*}_{t_L}(X)=i) P(A=a)$$
            
        For the case $i=\alpha$, observe that $I^{*}_{t_L}(X)=\alpha$ if and only if $x\notin dom(T_{t_L})$,
        which means if and only if $dom_{T_{t_L}}(X)=0$. But remembering that $dom(T_{t_L})$ and $dom(T_{t_R})$ form a partition of $dom(T_t)$, we have that $dom_{T_{t_L}}(X)=0$ if and only if $dom_{T_{t_R}}(X)=1$ or $dom_{T_t}(X)=0$. And as these two events are mutually exclusive and independent of $A$, by the part a) and the recursion hypothesis, we have:
            \begin{flalign*}
                 P(I^{*}_{t_L}(X)=\alpha, A=a) =\\
                 P\left( \left[dom_{T_{t_R}}(X)=1 \wedge A=a \right] \right . \vee \\
                 \left . \left[dom_{T_t}(X)=0 \wedge A=a \right] \right) =\\
                 P\left( dom_{T_{t_R}}(X)=1 \wedge A=a \right) + \\
                 P\left( dom_{T_t}(X)=0, A=a \right) = \\
                 P\left( dom_{T_{t_R}}(X)=1 \right) P\left( A=a \right) + \\
                 P\left( dom_{T_t}(X)=0\right) P\left( A=a \right) = \\
                 \left[P\left( dom_{T_{t_R}}(X)=1 \right) + P\left( dom_{T_t}(X)=0\right)\right] \\
                 P\left( A=a \right) = \\
                 P( I^{*}_{t_L}(X)=\alpha) P\left( A=a \right)
            \end{flalign*}

\end{itemize}
\end{proof}

\end{appendices}

\printbibliography[title={References}]
\end{document}